\documentclass[11pt,letterpaper]{article}
\usepackage{emnlp2017}
\usepackage{times}
\usepackage{latexsym}
\usepackage{array}
\usepackage{multirow}
\usepackage{amssymb}
\usepackage{amsmath}
\usepackage{mathtools}
\usepackage{amsthm}
\usepackage{xspace}
\usepackage{algorithm}
\usepackage{tikz-dependency}
\usepackage{cancel}

\usepackage[T1]{fontenc}
\usepackage[font=small,labelfont=bf,tableposition=top]{caption}
\DeclareCaptionLabelFormat{andtable}{#1~#2  \&  \tablename~\thetable}

\usepackage[T1]{fontenc}
\usepackage{lmodern}
\usepackage[utf8]{inputenc}

\usepackage{tabu}
\usepackage{xcolor}

\usepackage{graphicx}
\usepackage{caption}
\usepackage{subcaption}

\emnlpfinalcopy

\newcommand{\notes}[1]{}

\theoremstyle{definition}

\theoremstyle{plain}

\newcommand{\ith}[1]{\ensuremath{i^{{th}}}}

\newcount\permx
\newcount\permy
\def\permdot#1#2{
\permx=#1 \advance\permx by-1
\permy=#2 \advance\permy by-1
\psframe[fillcolor=black, fillstyle=solid]
(\permx,\permy)(#1, #2)
}

\newcommand{\boxnum}[1]{{\setlength{\fboxsep}{1pt}\raisebox{1pt}{\hspace{1pt}\fbox{\tiny #1}\hspace{1pt}}}}
\newcommand{\ind}[1]{\ensuremath{_{\kern-0.5pt\boxnum{#1}}}}

\newcommand{\vecI}{\ensuremath{\mathbf{I}}\xspace}

\newcommand{\smallnt}[1]{\ensuremath{_{\mbox{\tiny PP}}}\xspace}

\newcommand{\pseudocode}{Algorithm}
\floatname{algorithm}{\pseudocode}

\iffalse

\else

\fi

\def\emnlppaperid{***}

\makeatletter
\def\blfootnote{\gdef\@thefnmark{}\@footnotetext}
\makeatother

\title{OSU Multimodal Machine Translation System Report}

\author{Mingbo Ma, Dapeng Li, Kai Zhao$^\dagger$ and Liang Huang \\
Department of EECS\\
Oregon State University\\
Corvallis, OR 97331, USA \\
  {\tt \{mam, lidap, zhaok, liang.huang\}@oregonstate.edu}}

\date{}

\begin{document}

\maketitle

\begin{abstract}
This paper describes Oregon State University's submissions to the shared 
WMT'17 task ``multimodal translation task I''. 
In this task, all the sentence pairs are image captions 
in different languages. 
The key difference between this task and conventional machine translation is that 
we have corresponding images as additional information 
for each sentence pair. 
In this paper, we introduce a simple but effective system which takes an image
shared between different languages, 
feeding it into the both encoding and decoding side. 
We report our system's performance for English-French and English-German with Flickr30K (in-domain) and MSCOCO (out-of-domain) datasets. 
Our system achieves the best performance in TER for English-German for MSCOCO dataset.
\end{abstract}

\blfootnote{$^\dagger$ Current address: Google Inc., 111 8th Avenue, New York, New York, USA.}

\section{Introduction}
\label{sec:intro}

Natural language generation (NLG) is one of 
the most important tasks in natural language processing (NLP). 
It can be applied to a lot of interesting applications such like 
machine translation, image captioning, question answering.
In recent years, Recurrent Neural Networks (RNNs) based approaches have shown promising performance in generating more fluent and meaningful sentences compared with conventional models such as rule-based model~\cite{Mirkovic}, corpus-based n-gram models~\cite{WenGKMSVY15} and
trainable generators~\cite{Stent:2004}.

More recently, attention-based encoder-decoder models~\cite{Bahdanau:14} have been proposed to provide the decoder more accurate 
alignments to generate more relevant words. 
The remarkable ability of attention mechanisms quickly update the state-of-the-art performance on variety of NLG tasks, such as machine translation~\cite{Luong:15}, image captioning~\cite{Kelvin:15,zhilin:16}, and text summarization~\cite{Rush:15,Nallapati:16}.

However, for multimodal translation~\cite{Elliott:15}, where we translate a caption from one language into another given
a corresponding image, we need to design a new model since the decoder needs to consider both language and images at the same time. 

This paper describes our participation in the WMT 2017 multimodal task 1. Our model feeds the image information to both the encoder and decoder, to ground their hidden representation within the same context of image during training. In this way, during testing time, the decoder would generate more relevant words given the context of both source sentence and image. 
\section{Model Description}
\label{sec:model}

For the neural-based machine translation model, the encoder needs to 
map sequence of word embeddings from the source side into another 
representation of the entire sequence using 
recurrent networks. Then, in the second stage, decoder generates one 
word at a time with considering global (sentence representation) 
and local information (weighted context) from source side. For 
simplicity, our proposed model is based on the attention-based 
encoder-decoder framework in~\cite{Luong:15}, refereed as ``Global attention''.

On the other hand, for the early work of neural-basic caption 
generation models~\cite{VinyalsTBE15}, the convolutional neural 
networks (CNN) generate the image features which feed into the 
decoder directly for generating the description. 

The first stage of the above two tasks both map the temporal and spatial 
information into a fixed dimensional vector which makes it feasible 
to utilize both information at the same time. 

\begin{figure}[ht]
\includegraphics[width=7.5cm]{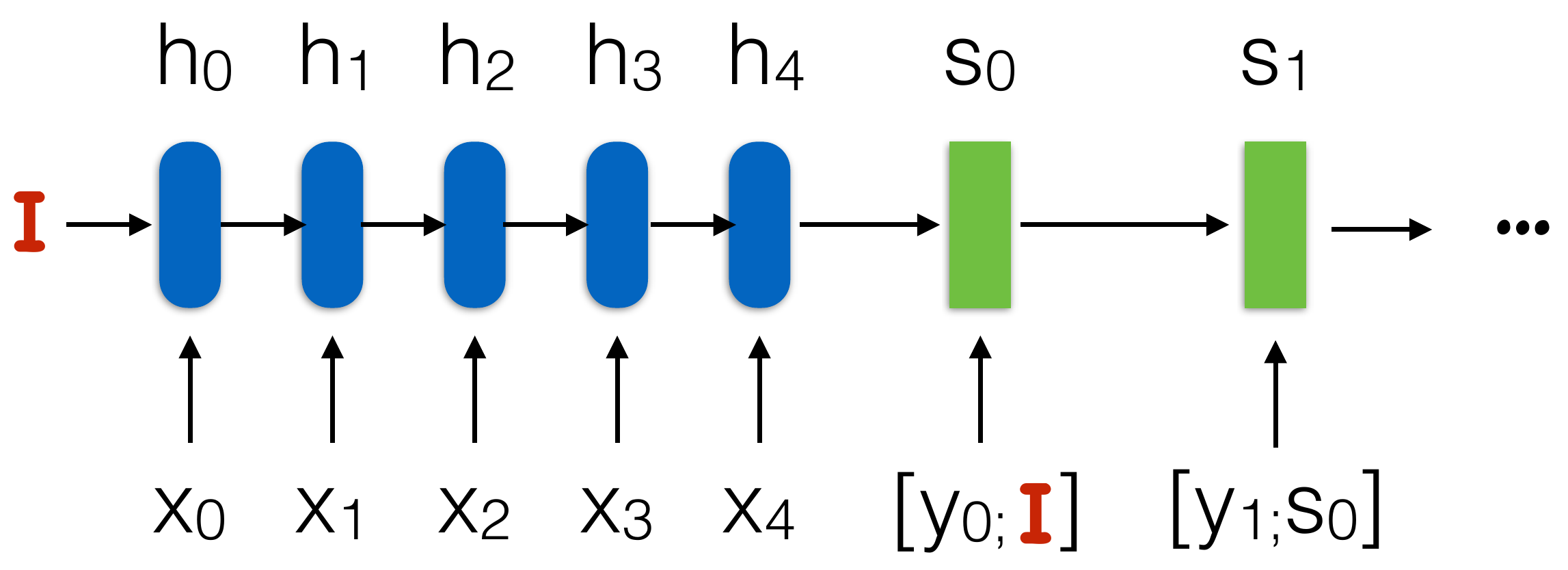}
\caption{The image information is feed to both encoder and decoder for initialization. I (in red) represents the image feature that are generated by CNN.}
\label{fig:modle}
\end{figure}

Fig.~\ref{fig:modle} shows the basic idea of our proposed model (OSU1). The red character $\vecI$ represents the image feature that is generated from CNN. In our case, we directly use the image features that are provided by WMT, and these features are generated by residual networks~\cite{kaiming:2016}.

The encoder (blue boxes) in Fig.~\ref{fig:modle} takes the image feature as initialization for generating
each hidden representation. This process is very similar to neural-basic caption 
generation~\cite{VinyalsTBE15} which grounds each word's hidden representation to 
the context given by the image. On the decoder side (green boxes in Fig.~\ref{fig:modle}), 
we not only let each decoded word align to source words by global attention but also 
feed the image feature as initialization to the decoder.

\section{Experiments}
\label{sec:exp}

\subsection{Datasets}

In our experiments, we use two datasets Flickr30K~\cite{elliott:2016} 
and MSCOCO~\cite{MSCOCO} which are provided by the WMT organization.
For both datasets, there are triples that contains English as source 
sentence, its German and French human translations and corresponding 
image. The system is only trained on Flickr30K datasets but are also tested on 
MSCOCO besides Flickr30K. 
MSCOCO datasets are considered out-of-domain (OOD) testing while 
Flickr30K dataset are considered in-domain testing.
The datasets' statics is shown in Table~\ref{tb:data}

\begin{table}[htbp]
\centering
\scalebox{0.9}{
\begin{tabular}{ |l|c|c|c|c| }
\hline
            Datasets     & Train     &  Dev &  Test  & OOD ?\\ 
\hline
            Flickr30K    &  $29,000$  &    $1,014$   & $1,000$   & No\\
\hline
            MSCOCO    &  -  &    -   & $461$   & Yes\\
\hline
\end{tabular}
}
\caption{Summary of datasets statistics.}
\label{tb:data}
\end{table}

\subsection{Training details}

For preprocessing, we convert all of the sentences to lower case, normalize the 
punctuation, and do the tokenization. For simplicity, our vocabulary keeps all the 
words that show in training set. 
For image representation, we use ResNet~\cite{kaiming:2016} generated image features which are
provided by the WMT organization. In our experiments, we only use average pooled features.

Our implementation is adapted from on Pytorch-based OpenNMT~\cite{2017opennmt}. We use two layered
bi-LSTM~\cite{Sutskever:2014} on the source side as encoder. Our batch size is 64, with SGD optimization and a learning rate at 1.
For English to German, the dropout rate 
is 0.6, and for English to French, the dropout rate is 0.4. These two parameters are selected by
observing the performance on development set. Our word embeddings are randomly initialized with 
500 dimensions. 
The source side vocabulary is 10,214 and the target side vocabulary is 18,726 for German 
and 11,222 for French.

\subsection{Beam search with length reward}

During test time, beam search is widely used to improve the output text quality by giving the decoder 
more options to generate the next possible word.
However, different from traditional beam search in phrase-based MT where all hypotheses 
know the number of steps to finish the generation, while in neural-based generation, 
there is no information about what is the most ideal number of steps to finish the decoding.
The above issue also leads to another problem that the beam search in neural-based MT prefers shorter sequences due to probability-based scores for evaluating different candidates.
In this paper, we use Optimal Beam Search~\cite{huang+:2017} (OBS) during decoding time.
OBS uses bounded length reward
mechanism which allows a modified version of
our beam search algorithm to remain optimal.

\begin{figure*}
\centering
\begin{minipage}{0.5\textwidth}
  \centering
  \includegraphics[width=1\linewidth]{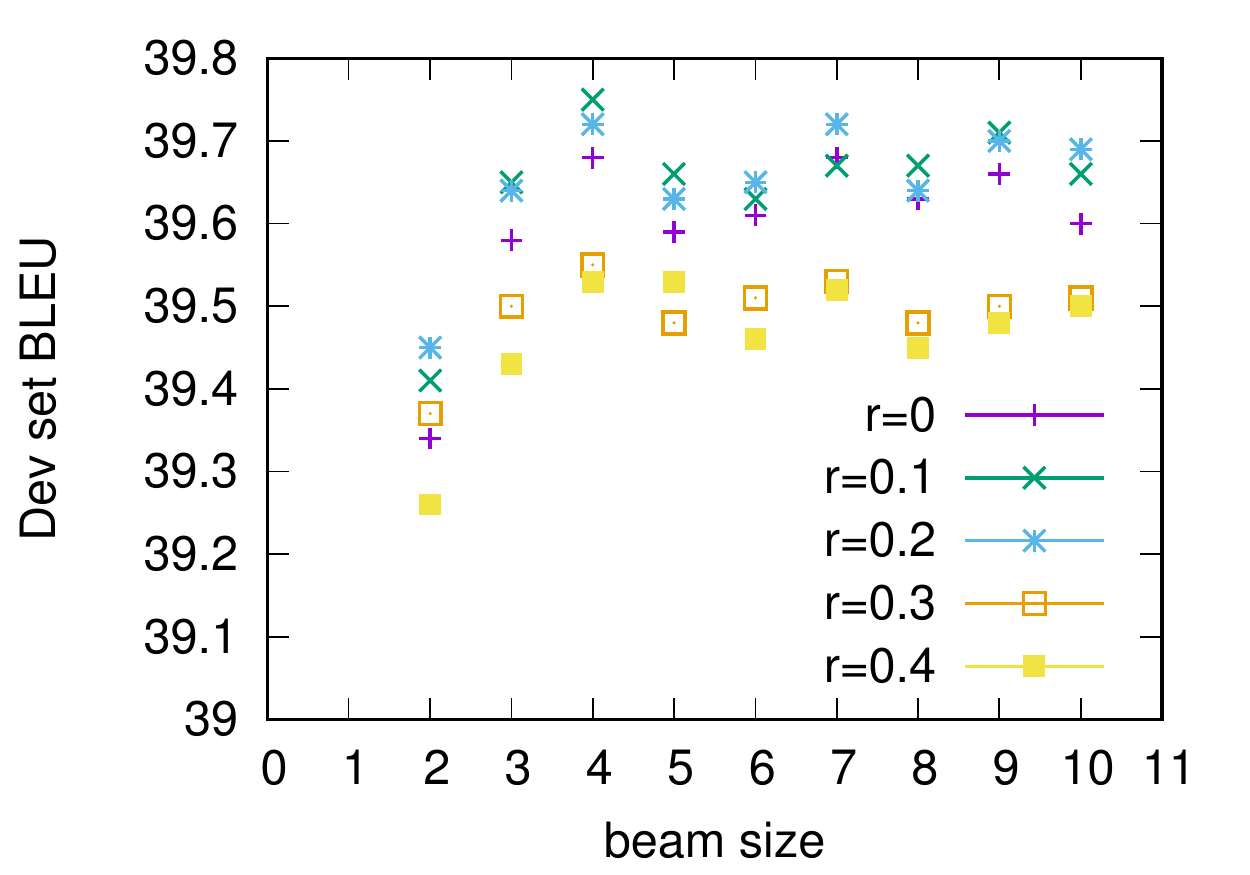}
  \captionof{figure}{BLEU vs. beam size}
  \label{fig:b}
\end{minipage}%
\begin{minipage}{.5\textwidth}
  \centering
  \includegraphics[width=1\linewidth]{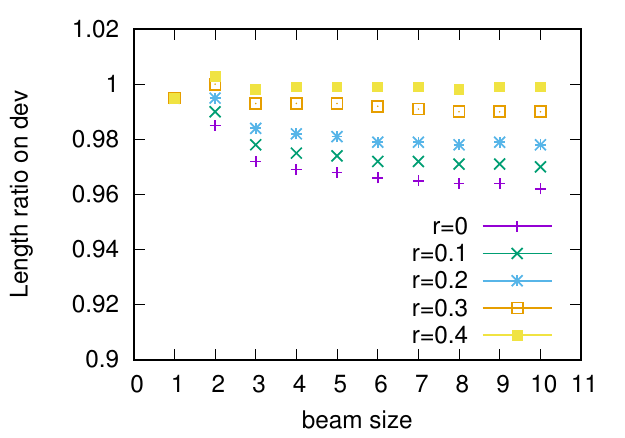}
  \captionof{figure}{length ratio vs. beam size}
  \label{fig:l}
\end{minipage}
\end{figure*}

Figure~\ref{fig:b} and Figure~\ref{fig:l} show the BLEU score and length ratio with different rewards for different beam size. We choose beam size equals to 5 and reward equals to 0.1 during decoding. 

\subsection{Results}

WMT organization provides three different evaluating metrics: BLEU~\cite{Papineni:2002}, METEOR~\cite{Lavie2009} and TER~\cite{Snover06astudy}. 

Table~\ref{tb:fen2de} to Table~\ref{tb:cen2fr} summarize the performance with their corresponding rank among all other systems. 
We only 
show a few top performing systems in the tables to make a comparison. OSU1 is our proposed model and OSU2 
is our baseline system without any image information. 
For MSCOCO dataset, the translation from English to 
German (Table~\ref{tb:cen2de}), which is the hardest tasks compared with others since it is from 
English to German on OOD dataset, we 
achieve best TER score across all other systems.

\begin{table}[htbp]
\centering
\scalebox{0.9}{
\begin{tabular}{ |c|c|c|c|c| }
\hline 
            System      & Rank & TER     &  METEOR &  BLEU  \\ 
\hline
            UvA-TiCC    & 1   &  \textbf{47.5}  &    53.5   & \textbf{33.3} \\
\hline
            NICT        & 2 &  48.1 &    \textbf{53.9}   & 31.9  \\
\hline
            LIUMCVC   & 3 \& 4 &  48.2 &    53.8   & 33.2  \\
\hline
            CUNI   & 5 &  50.7 &    51   & 31.1  \\
\hline
            $\text{OSU2}^\dagger$  & 6 &  50.7 &   50.6   & 31  \\
\hline
            $\text{OSU1}^\dagger$  & 8 &  51.6 &   48.9   & 29.7  \\
\hline
\end{tabular}
}
\caption{Experiments on Flickr30K dataset for translation from English to German. 16 systems in total. $\dagger$ represents our system.}
\label{tb:fen2de}
\end{table}

\begin{table}[htbp]
\centering
\scalebox{0.9}{
\begin{tabular}{ |c|c|c|c|c| }
\hline
            System    & Rank & TER     &  METEOR &  BLEU  \\ 
\hline
            \color{blue}\text{OSU1}   & \color{blue}1 &  \color{blue}\textbf{52.3}  &    \color{blue}46.5   & \color{blue}27.4 \\
\hline
            UvA-TiCC   & 2 &  52.4 &    48.1   & 28  \\
\hline
            LIUMCVC   &  3 &  52.5 &    \textbf{48.9}   & \textbf{28.7}  \\
\hline
            \color{blue}$\text{OSU2}$  & \color{blue}8 &  \color{blue}55.9 &   \color{blue}45.7  & \color{blue}26.1  \\
\hline
\end{tabular}
}
\caption{Experiments on MSCOCO dataset for translation from English to German. 15 systems in total. $\dagger$ represents our system.}
\label{tb:cen2de}
\end{table}

\begin{table}[htbp]
\centering
\scalebox{0.9}{
\begin{tabular}{ |c|c|c|c|c| }
\hline
            System   & Rank  & TER     &  METEOR &  BLEU  \\ 
\hline
            LIUMCVC  & 1  &  \textbf{28.4}  &   \textbf{72.1}   & \textbf{55.9} \\
\hline
            NICT  & 2  &  28.4 &    72   & 55.3  \\
\hline
            DCU   & 3  &  30 &    70.1   & 54.1  \\
\hline
            $\text{OSU2}^\dagger$ & 5   &  32.7 &   68.3   & 51.9 \\
\hline
            $\text{OSU1}^\dagger$ & 6    &  33.6 &   67.2   & 51 \\
\hline
\end{tabular}
}
\caption{Experiments on Flickr30K dataset for translation from English to French. 11 systems in total. $\dagger$ represents our system.}
\label{tb:fen2fr}
\end{table}

\begin{table}[htbp]
\centering
\scalebox{0.9}{
\begin{tabular}{ |c|c|c|c|c| }
\hline
            System    & Rank & TER     &  METEOR &  BLEU  \\ 
\hline
            LIUMCVC  & 1  &  \textbf{34.2}  &   \textbf{65.9}   & \textbf{45.9} \\
\hline
            NICT   & 2   &  34.7 &    65.6   & 45.1  \\
\hline
            DCU  & 3   &  35.2 &    64.1   & 44.5 \\
\hline
            \color{blue}$\text{OSU2}$  & \color{blue}4  &  \color{blue}36.7 &   \color{blue}63.8   & \color{blue}44.1 \\
\hline
            \color{blue}$\text{OSU1}$  & \color{blue}6  &  \color{blue}37.8 &   \color{blue}61.6   & \color{blue}41.2 \\
\hline
\end{tabular}
}
\caption{Experiments on MSCOCO dataset for translation from English to French. 11 systems in total.}
\label{tb:cen2fr}
\end{table}

\begin{figure*}[htbp]
\centering
   \begin{subfigure}[b]{1\textwidth}
   \includegraphics[width=2.5cm]{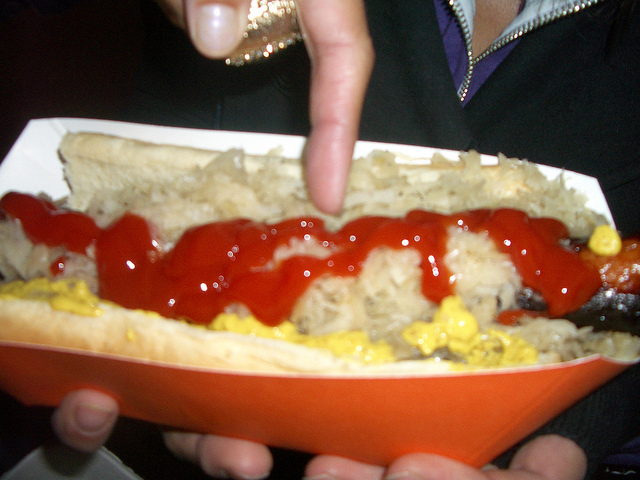}
   \;
    \scalebox{0.85}{
    \begin{tabular}[b]{cl}\hline
      input & a finger pointing at a hotdog with cheese , sauerkraut and ketchup . \\
      OSU1 &  ein finger zeigt auf einen hot dog mit einem messer , wischmobs und napa . \\
      OSU2 &  ein finger zeigt auf einen hotdog mit hammer und italien .  \\
      Reference & ein finger zeigt auf einen hotdog mit käse , sauerkraut und ketchup .  \\ \hline
    \end{tabular}}
\end{subfigure}

\vspace{3mm}

\begin{subfigure}[b]{1\textwidth}
   \includegraphics[width=2.5cm]{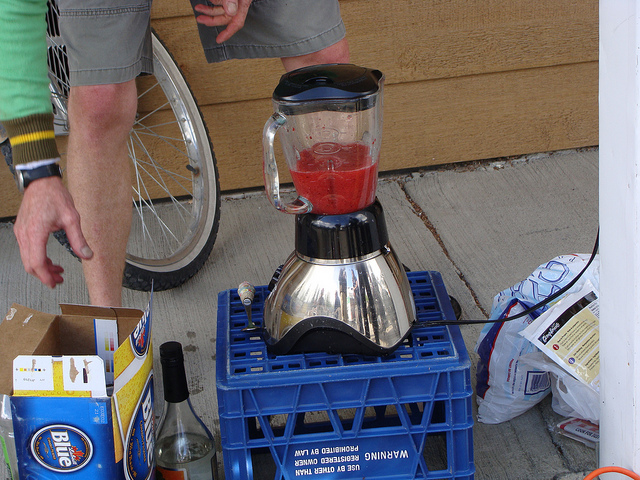}
   \;
    \scalebox{0.85}{
    \begin{tabular}[b]{cl}\hline
      input & a man reaching down for something in a box  \\
      OSU1 &   ein mann greift nach unten , um etwas zu irgendeinem .  \\
      OSU2 &  ein mann greift nach etwas in einer kiste .   \\
      Reference & ein mann bückt sich nach etwas in einer schachtel .   \\ \hline
    \end{tabular}}
\end{subfigure}
\caption{Two testing examples that image information confuses the NMT model.}
\label{tb:bad}
\end{figure*}

\begin{figure*}[htbp]
\centering
   \begin{subfigure}[b]{1\textwidth}
   \includegraphics[width=2.5cm]{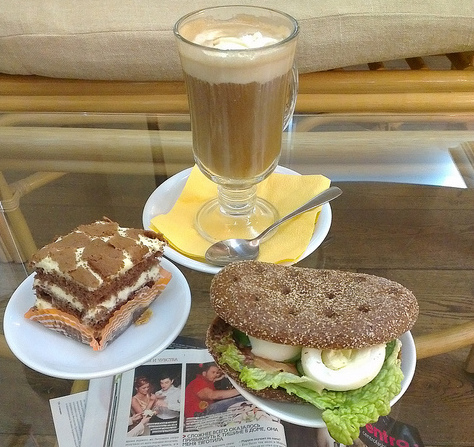}
   \;
    \scalebox{0.85}{
    \begin{tabular}[b]{cl}\hline
      input & there are two foods and one drink set on the clear table .  \\
      OSU1 & da sind zwei speisen und ein getränk am klaren tisch .   \\
      OSU2 &  zwei erwachsene und ein erwachsener befinden sich auf dem rechteckigen tisch .    \\
      Reference & auf dem transparenten tisch stehen zwei speisen und ein getränk .    \\ \hline
    \end{tabular}}
\end{subfigure}

\vspace{3mm}

\begin{subfigure}[b]{1\textwidth}
   \includegraphics[width=2.5cm]{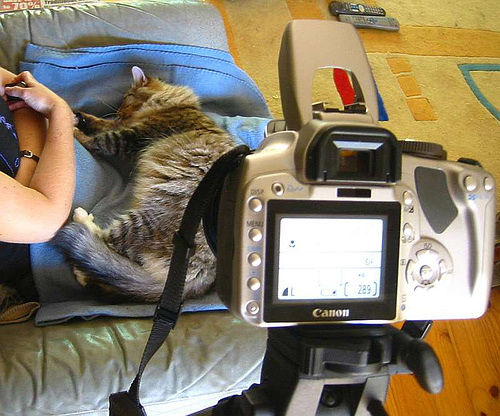}
   \;
    \scalebox{0.85}{
    \begin{tabular}[b]{cl}\hline
      input & a camera set up in front of a sleeping cat .   \\
      OSU1 & eine kameracrew vor einer schlafenden katze .     \\
      OSU2 &  eine kamera vor einer blonden katze .     \\
      Reference & eine kamera , die vor einer schlafenden katze aufgebaut ist     \\ \hline
    \end{tabular}}
\end{subfigure}
\caption{Two testing examples that image information helps the NMT model.}
\label{tb:good}
\end{figure*}

As describe in section~\ref{sec:model}, OSU1 is the model with image information for both encoder 
and decoder, and OSU2 is only the neural machine translation baseline without any image information. 
From the above results table we found that image information would hurt the performance in some cases.
In order to have more detailed analysis, we show some test examples for the translation from English to 
German on MSCOCO dataset. 

Fig~\ref{tb:bad} shows two examples that NMT baseline model performances better than OSU1 model. 
In the first example, OSU1 generates several unseen objects from given image, such like knife. 
The image feature might not represent the image accurately.
For the second example, OSU1 model ignores the object ``box'' in the image.

Fig~\ref{tb:good} shows two examples that image feature helps the OSU1 to generate better results.
In the first example, image feature successfully detects the object ``drink'' while the baseline
completely neglects this.
In the second example, the image feature even help the model figure out the action of the cat is ``sleeping''.

\section{Conclusion}
We describe our system submission to the shared 
WMT'17 task ``multimodal translation task I''.
The results for English-German and English-French on Flickr30K and MSCOCO datasets are reported 
in this paper.
Our proposed model is simple but effective and we achieve the best performance in TER for 
English-German for MSCOCO dataset. 

\section{Acknowledgment}
This work is supported in part by 
NSF IIS-1656051,
DARPA FA8750-13-2-0041 (DEFT),
DARPA N66001-17-2-4030 (XAI),
a Google Faculty Research Award,
and an HP Gift.

\bibliography{acl2017}

\begin{thebibliography}{}
\expandafter\ifx\csname natexlab\endcsname\relax\def\natexlab#1{#1}\fi

\bibitem[{Bahdanau et~al.(2014)Bahdanau, Cho, and Bengio}]{Bahdanau:14}
Dzmitry Bahdanau, Kyunghyun Cho, and Yoshua Bengio. 2014.
\newblock Neural machine translation by jointly learning to align and
  translate.
\newblock {\em CoRR\/} .

\bibitem[{{Elliott} et~al.(2016){Elliott}, {Frank}, {Sima'an}, and
  {Specia}}]{elliott:2016}
D.~{Elliott}, S.~{Frank}, K.~{Sima'an}, and L.~{Specia}. 2016.
\newblock Multi30k: Multilingual english-german image descriptions.
\newblock {\em Proceedings of the 5th Workshop on Vision and Language\/} pages
  70--74.

\bibitem[{Elliott et~al.(2015)Elliott, Frank, and Hasler}]{Elliott:15}
Desmond Elliott, Stella Frank, and Eva Hasler. 2015.
\newblock Multi-language image description with neural sequence models.
\newblock {\em CoRR\/} .

\bibitem[{He et~al.(2016)He, Zhang, Ren, and Sun}]{kaiming:2016}
Kaiming He, Xiangyu Zhang, Shaoqing Ren, and Jian Sun. 2016.
\newblock Deep residual learning for image recognition.
\newblock {\em Conference on Computer Vision and Pattern Recognition {CVPR}\/}
  .

\bibitem[{Huang et~al.(2017)Huang, Zhao, and Ma}]{huang+:2017}
Liang Huang, Kai Zhao, and Mingbo Ma. 2017.
\newblock When to finish? optimal beam search for neural text generation
  (modulo beam size).
\newblock In {\em EMNLP 2017\/}.

\bibitem[{{Klein} et~al.(2017){Klein}, {Kim}, {Deng}, {Senellart}, and
  {Rush}}]{2017opennmt}
G.~{Klein}, Y.~{Kim}, Y.~{Deng}, J.~{Senellart}, and A.~M. {Rush}. 2017.
\newblock Opennmt: Open-source toolkit for neural machine translation.
\newblock {\em ArXiv e-prints\/} .

\bibitem[{Lavie and Denkowski(2009)}]{Lavie2009}
Alon Lavie and Michael~J. Denkowski. 2009.
\newblock The meteor metric for automatic evaluation of machine translation.
\newblock {\em Machine Translation\/} .

\bibitem[{Lin et~al.(2014)Lin, Maire, Belongie, Bourdev, Girshick, Hays,
  Perona, Ramanan, Doll{\'{a}}r, and Zitnick}]{MSCOCO}
Tsung{-}Yi Lin, Michael Maire, Serge~J. Belongie, Lubomir~D. Bourdev, Ross~B.
  Girshick, James Hays, Pietro Perona, Deva Ramanan, Piotr Doll{\'{a}}r, and
  C.~Lawrence Zitnick. 2014.
\newblock Microsoft {COCO:} common objects in context .

\bibitem[{Luong et~al.(2015)Luong, Pham, and Manning}]{Luong:15}
Minh-Thang Luong, Hieu Pham, and Christopher~D. Manning. 2015.
\newblock Effective approaches to attention-based neural machine translation.
\newblock {\em CoRR\/} .

\bibitem[{Mirkovic et~al.(2011)Mirkovic, Cavedon, Purver, Ratiu, Scheideck,
  Weng, Zhang, and Xu}]{Mirkovic}
Danilo Mirkovic, Lawrence Cavedon, Matthew Purver, Florin Ratiu, Tobias
  Scheideck, Fuliang Weng, Qi~Zhang, and Kui Xu. 2011.
\newblock Dialogue management using scripts and combined confidence scores.
\newblock {\em US Patent\/} pages 7,904,297.

\bibitem[{Nallapati et~al.(2016)Nallapati, Zhou, and Ma}]{Nallapati:16}
Ramesh Nallapati, Bowen Zhou, and Mingbo Ma. 2016.
\newblock Classify or select: Neural architectures for extractive document
  summarization.
\newblock {\em CoRR\/} .

\bibitem[{Papineni et~al.(2002)Papineni, Roukos, Ward, and Zhu}]{Papineni:2002}
Kishore Papineni, Salim Roukos, Todd Ward, and Wei-Jing Zhu. 2002.
\newblock Bleu: A method for automatic evaluation of machine translation.
\newblock {\em Proceedings of the 40th Annual Meeting on Association for
  Computational Linguistics\/} .

\bibitem[{Rush et~al.(2015)Rush, Chopra, and Weston}]{Rush:15}
Alexander~M. Rush, Sumit Chopra, and Jason Weston. 2015.
\newblock A neural attention model for abstractive sentence summarization .

\bibitem[{Snover et~al.(2006)Snover, Dorr, Schwartz, Micciulla, and
  Makhoul}]{Snover06astudy}
Matthew Snover, Bonnie Dorr, Richard Schwartz, Linnea Micciulla, and John
  Makhoul. 2006.
\newblock A study of translation edit rate with targeted human annotation.
\newblock {\em In Proceedings of Association for Machine Translation in the
  Americas\/} .

\bibitem[{Stent et~al.(2004)Stent, Prasad, and Walker}]{Stent:2004}
Amanda Stent, Rashmi Prasad, and Marilyn Walker. 2004.
\newblock Trainable sentence planning for complex information presentation in
  spoken dialog systems.
\newblock {\em Proceedings of the 42Nd Annual Meeting on Association for
  Computational Linguistics\/} .

\bibitem[{Sutskever et~al.(2014)Sutskever, Vinyals, and Le}]{Sutskever:2014}
Ilya Sutskever, Oriol Vinyals, and Quoc~V. Le. 2014.
\newblock Sequence to sequence learning with neural networks.
\newblock {\em Proceedings of the 27th International Conference on Neural
  Information Processing Systems\/} .

\bibitem[{Vinyals et~al.(2015)Vinyals, Toshev, Bengio, and
  Erhan}]{VinyalsTBE15}
Oriol Vinyals, Alexander Toshev, Samy Bengio, and Dumitru Erhan. 2015.
\newblock Show and tell: {A} neural image caption generator.
\newblock {\em {IEEE} Conference on Computer Vision and Pattern Recognition\/}
  pages 3156--3164.

\bibitem[{Wen et~al.(2015)Wen, Gasic, Kim, Mrksic, Su, Vandyke, and
  Young}]{WenGKMSVY15}
Tsung{-}Hsien Wen, Milica Gasic, Dongho Kim, Nikola Mrksic, Pei{-}hao Su, David
  Vandyke, and Steve~J. Young. 2015.
\newblock Stochastic language generation in dialogue using recurrent neural
  networks with convolutional sentence reranking.
\newblock {\em CoRR\/} .

\bibitem[{Xu et~al.(2015)Xu, Ba, Kiros, Cho, Courville, Salakhudinov, Zemel,
  and Bengio}]{Kelvin:15}
Kelvin Xu, Jimmy Ba, Ryan Kiros, Kyunghyun Cho, Aaron Courville, Ruslan
  Salakhudinov, Rich Zemel, and Yoshua Bengio. 2015.
\newblock Show, attend and tell: Neural image caption generation with visual
  attention.
\newblock {\em Proceedings of the 32nd International Conference on Machine
  Learning (ICML-15)\/} .

\bibitem[{Yang et~al.(2016)Yang, Yuan, Wu, Cohen, and
  Salakhutdinov}]{zhilin:16}
Zhilin Yang, Ye~Yuan, Yuexin Wu, William~W. Cohen, and Ruslan Salakhutdinov.
  2016.
\newblock Review networks for caption generation.
\newblock {\em Advances in Neural Information Processing Systems\/} .

\end{thebibliography}
\bibliographystyle{emnlp_natbib}

\end{document}